\documentclass[11pt,a4paper]{article}
\usepackage[hyperref]{emnlp2021}
\usepackage{times}
\usepackage{latexsym}

\usepackage[utf8]{inputenc} 
\usepackage{microtype}
\usepackage{graphicx}
\usepackage{subfigure}
\usepackage{booktabs} 
\usepackage{comment}
\usepackage{hyperref}
\usepackage{mathtools}
\usepackage{color}
\usepackage{graphicx}
\usepackage{CJKutf8}
\usepackage{url}
\usepackage{multicol,multirow}
\usepackage{arydshln}
\usepackage{makecell}
\usepackage{array}
\usepackage{bm}
\usepackage{cancel}
\usepackage{parskip}
\usepackage[T1]{fontenc}    
\usepackage{booktabs}       
\usepackage{amsmath, amsthm}
\usepackage{amssymb}
\usepackage{amsfonts}
\usepackage[ruled]{algorithm2e}

\usepackage{enumitem}

\DeclareMathOperator*{\argmax}{arg\,max}

\usepackage{microtype}
\definecolor{ao}{rgb}{0.0, 0.5, 0.0}
\definecolor{asparagus}{rgb}{0.53, 0.66, 0.42}
\definecolor{amber}{rgb}{1.0, 0.49, 0.0}
\definecolor{alizarin}{rgb}{0.82, 0.1, 0.26}
\definecolor{applegreen}{rgb}{0.55, 0.71, 0.0}
\definecolor{amethyst}{rgb}{0.6, 0.4, 0.8}
\definecolor{auburn}{rgb}{0.43, 0.21, 0.1}
\def\aclpaperid{1198} 

\title{Layer-wise Model Pruning based on Mutual Information}
\date{}

\author{
Chun Fan$^{\spadesuit\bigstar}$, Jiwei Li$^{\blacklozenge\clubsuit}$, Xiang Ao$^\blacktriangledown$,\\
{\bf Fei Wu}$^\blacklozenge$, {\bf Yuxian Meng}$^\clubsuit$ and {\bf Xiaofei Sun}$^\clubsuit$ \\
  $^\blacklozenge$Zhejiang University\\
  $^\spadesuit$Computer Center of Peking University,
  $^\bigstar$Peng Cheng Laboratory\\
  $^\blacktriangledown$Key Lab of Intelligent Information Processing of Chinese Academy of Sciences\\
  $^\clubsuit$ Shannon.AI\\
  \{yuxian\_meng, xiaofei\_sun, jiwei\_li\}@shannonai.com\\
  fanchun@pku.edu.cn,
  aoxiang@ict.ac.cn,
  wufei@zju.edu.cn
}

\begin{document}
\maketitle

\begin{abstract}
Inspired by mutual information (MI) based feature selection in SVMs and logistic regression, in this paper, we propose MI-based layer-wise pruning: for each layer of a multi-layer neural network, neurons with higher values of MI with respect to preserved neurons in the upper   layer are preserved. Starting from the top softmax layer, layer-wise pruning proceeds in a top-down fashion until reaching the bottom word embedding layer. 
The proposed pruning strategy offers  merits over weight-based pruning techniques: (1) it avoids irregular memory access since representations  and matrices can be squeezed into their smaller but dense counterparts, leading to  greater speedup; (2) in a manner of top-down pruning, the proposed method operates from a more global perspective based on training signals in the top layer, and prunes each layer by propagating the effect of global signals through  layers, leading to better performances at the same sparsity level. 
Extensive experiments show that at the same sparsity level, the proposed  strategy offers both greater speedup and higher performances than  weight-based pruning methods (e.g., magnitude pruning, movement pruning).\footnote{To appear at EMNLP21.} 
\end{abstract}

\section{Introduction}
In spite of  impressive results of neural networks, the huge model size has hindered their applications in cases where computation and memory resources are limited.\footnote{For example, the  GPT-3 model \citep{brown2020language} has 175B parameters in total, with 96 layers and 96 attention heads \citep{vaswani2017attention} per layer.} As a result, training and using existing huge models not only requires rich hardware resources, but also consumes high environmental costs \citep{strubell2019energy}.

Model pruning,   reduces model sizes by dropping a fraction of the model parameters, 
 to reduce computation intensity and memory footprint of large models 
at the lowest cost of accuracy on end tasks
 \citep{joulin2016fasttext,ganesh2020compressing,gordon-etal-2020-compressing}. Among pruning techniques, weight based pruning is a widely-used group of methods. 
 It focuses on removing weights according to their importance under different specific criteria, e.g., the magnitude \citep{han2015learning, han2015deep}, first-order derivative \citep{lee2018snip,sanh2020movement} and second-order derivative information \citep{NIPS1989_6c9882bb,NIPS1992_303ed4c6}, and it has been successfully applied to a large variety of model architectures \citep{guo2016dynamic,gale2019state,Molchanov_2019_CVPR} and downstream tasks \citep{mccarley2019pruning,gordon-etal-2020-compressing}. 

While weight-based  methods have been successfully applied to a wide range of neural models
for model pruning, 
they come with the following shortcomings: 
(1) weights in matrices are pruned irregularly, which lead to irregular memory access, resulting in runtime inefficiency;
 (2)  weight matrices are pruned independently, and this neglect of 
 global supervision from training signals at the top layer
    and ignorance of
 information 
 propagation   between consecutive  layers
  may result in sub-optimality of pruned networks.

In this paper, inspired by mutual information (MI) based feature selection \citep{kuncheva2007stability} in  
SVMs and logistic regression, we propose 
MI based
 layer-wise pruning, 
 to address the aforementioned drawbacks of weight-based pruning methods in NLP. 
  For each layer of a multi-layer neural network,  neurons
  with higher values of MI 
with respect   to the preserved neurons in the upper   
    layer are 
  preserved. Starting from the top softmax layer, layer-wise pruning proceeds until 
    reaching the bottom 
     input word embedding layer in a top-down fashion. 
          Once the preserved neurons in each layer are selected, the redundant  dimensions along with the corresponding rows and columns of the weight matrices can be pruned or squeezed, inducing model sparsity at different levels.
          
     The proposed  pruning strategy naturally addresses the aforementioned two shortcomings of weight-based methods: 
     (1) it avoids irregular memory access 
   since it   
      squeezes the pruned representations  and matrices into their smaller but dense counterparts.
      This 
       enables significantly faster computations than  weight-based pruning methods at the same sparsity level; (2) 
       rather than viewing each weight matrix  separately based on their own weight values, 
      the proposed method operates from a more global perspective 
   based on training signals at the top layer, 
       and prunes each layer by propagating 
      the effect of global 
        training signals through consecutive layers in a top-down fashion.
This leads to better performances at the same sparsity level.

We conduct extensive experiments on both generative tasks (MT) and discriminative tasks (question answering) in NLP
 to examine the effectiveness of the proposed  strategy. 
 We show that compared to weight-based pruning methods including magnitude pruning \citep{han2015learning}, movement pruning \citep{sanh2020movement} and $L_0$ pruning \citep{louizos2017learning}, the proposed method yields greater speedup along with better performances for the same sparsity levels on
 generative NLP tasks of WMT'14 En$\rightarrow$Fr and WMT'14 En$\rightarrow$De, 
 and discriminative NLP tasks of SQuAD v1.1 \citep{rajpurkar2016squad}, MNIL \cite{williams2017broad} and SST-5 \cite{socher2013recursive}. 
 In addition, we also show that the proposed method serves the feature selection purposes, where we observe significant performance boosts when fixing preserved neurons and relearning the pruned ones, leading to a state-of-the-art performance of 43.9 BLEU score for En$\rightarrow$Fr translation in setups without back-translation or external data. 

\section{Related Work}
\subsection{Model Pruning}
\paragraph{Generic Model Pruning}
Model pruning refers to reducing the model size by dropping a fraction of the model parameters, which dates back to early works of Optimal Brain Damage (PBD) \citep{NIPS1989_6c9882bb} and Optimal Brain Surgeon (OBS) \citep{NIPS1992_303ed4c6}. 
One major branch of neural model pruning methods is magnitude pruning \citep{han2015learning,see2016compression,narang2017exploring,Molchanov_2019_CVPR,gale2019state,frankle2020linear}, which  prunes model parameters measured by their importance scores. \newcite{han2015learning} removed all parameters with weight values below a threshold, and then retrained the remaining sparse network. 
\citet{guo2016dynamic} proposed {\it dynamic network surgery}, 
allowing for model connection recovery from incorrect pruning decisions made in previous iterations. 
\citet{zhu2018to} adopted a gradual pruning schedule, in which the sparsity level increases from an initial sparsity value to a specified final sparsity value during training. 
Other methods for neural model pruning include $L_0$ regularization pruning \citep{louizos2017learning}, variational dropout pruning \citep{kingma2015variational,molchanov2017variational,gomez2019learning} and movement pruning \citep{sanh2020movement}, etc.
Recent works have proposed a line of techniques to prune and produce sparsity in a structured way \citep{anwar2017structured,zhou2016less,hu2016network,liu2018rethinking}, which aims at pruning full convolutional filters or whole layers. Methods for structured pruning mainly include group Lasso \citep{NIPS2016_6e7d2da6,wen2016learning,he2017channel}, sparsity regularization \citep{li2016pruning,liu2017learning,huang2018data,gordon2018morphnet} and automatic network searching \citep{he2018amc,yu2019autoslim,NEURIPS2019_a01a0380,pmlr-v97-ding19a}.

\paragraph{Pruning Transformers}
Pruning Transformer based models has been of growing interest \citep{guo2019reweighted,chen2020lottery,li2020sac}.
\citet{fan2019reducing} proposed {\it LayerDrop} to reduce Transformer depth. 
\citet{NEURIPS2019_2c601ad9} 
proposed to use {\it head importance score} to prune BERT attention heads.
Attention heads can also be pruned by using $L_0$ regularization \citep{voita-etal-2019-analyzing} and cascade pruning \citep{wang2021spatten}.
\citet{wang-etal-2020-structured} combined $L_0$ regularization with matrix factorization to prune BERT. 
\citet{gordon-etal-2020-compressing} 
proposed that BERT can be pruned once during pre-training rather than separately for each task without sacrificing performance.

\subsection{Mutual Information Feature Selection}
Feature selection is the process of selecting a proper subset of features for better model performances
\citep{kira1992practical,guyon2003introduction,chandrashekar2014survey,bolon2016feature,cai2018feature}. A widely used method for feature selection is {\it Mutual Information Based Feature Selection} \citep{vergara2014review,liu2009feature,beraha2019feature}, which selects features that minimize the redundancy and maximize the relevance w.r.t. the target variable.
Various approaches including minimum-Redundancy-Maximum-Relevance (mRMR)  \citep{estevez2009normalized,brown2012conditional,bennasar2015feature} are proposed to accurately select features.

\section{Model}
\subsection{Overview for Model Pruning}
Given a set of inputs $\mathcal{M} = \{(X, Y)\}$, where 
each  input is a  word sequence $X = \{x_1, ..., x_t, ...,x_{N_x}\}$ and $N_x$ denotes the length of the input, 
our goal is to predict the label(s) for $X$, denoted by $Y$.

In a standard multi-layer neural network setup, the input layer first maps each input word $x_t$ to a vector representation $h_t^0\in\mathbb{R}^{D\times 1}$, where $D$ denotes the dimensionality. 
On top of the input layer, the model stacks
L intermediate neural layers. 
Let $h_t^l\in\mathbb{R}^{D\times 1}$  denote the representation for token $x_t$ at the $l^{\text{th}}$ layer.
$H^l\in\mathbb{R}^{D\times N}$ is the concatenation of representations at the $l^{\text{th}}$ layer for all tokens in the input $X$.
Each layer of the network involves multiple operations such as fully connected operations, ReLU, self-attentions  or residual connections.
The group of all operations within layer $l$ is
denoted by 
$F_l$, which maps $H^{l}$ to $H^{l+1}$: 
\begin{equation}
H^{l+1} = F_l(H^{l})
\label{layer}
\end{equation}
 The output from the last layer $h_t^L$ is fed to the final softmax layer for predictions. 
  To prune a neural network model, 
  let $m^l\in \{0,1\}^{D\times 1}$ denote the mask for representation dimensions at layer $l$.
  The number of {\bf $1$}s  in $m^l$ is a pre-defined hyper-parameter, denoted by $K$, controlling the sparsity of the network.  
$M^l\in \{0,1\}^{D\times N}$ makes $N$ copies of $m^l$, making the dimensionality of the mask the same as that of layer representations for $X$. 
Let $u^l$ denote the set of indexes for preserved dimensions, where $m^l [j\;\text{for}\;j\;\text{in}\;u^l] =1$. Eq.(\ref{layer}) can be rewritten as: 
\begin{equation}
H^{l+1} = F_l(H^{l} \otimes M^l)
\label{Hadamard}
\end{equation}
 where $\otimes$
  is the Hadamard
product. 
We need special attentions for the uppermost softmax layer.  No dimension should be pruned for
this layer since each dimension corresponds to an output label. 
$m^{\text{softmax}}= [1]^{|\mathcal{Y}|}$, where $|\mathcal{Y}|$ denotes the size of the output label set.

\subsection{Layer-wise Pruning}
\begin{figure}[t]
  \centering
  \includegraphics[scale=0.3]{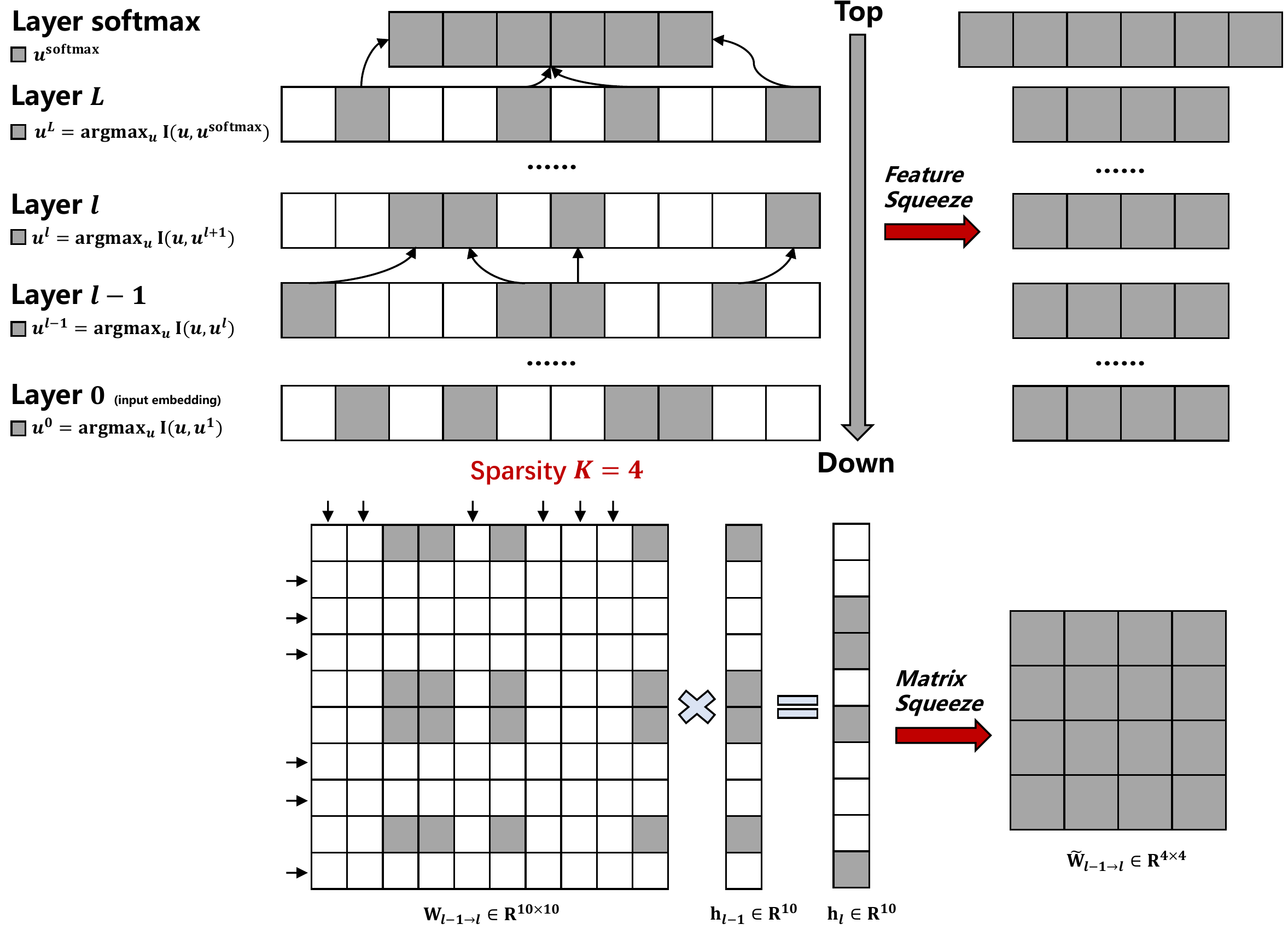}
  \caption{An overview of the proposed layer-wise pruning method. The top part shows pruning at the feature level, and the bottom part shows the weight matrix level pruning. Layer-wise pruning first selects feature dimensions in each layer regarding some correlation criterion $I(\cdot,\cdot)$, and then prunes matrix rows and cols according to the selected dimensions at consecutive layers, after which both features and matrices can be squeezed.}
  \label{fig:overview}
\end{figure}

The key point of layer-wise pruning is to construct correlations between dimensions in two consecutive layers $l-1$ and $l$. 
Then based on the correlations, we can prune the network in a top-down fashion:
with respect to output labels in the final softmax layer, we select  the top $K$ correlated dimensions in the $L^\text{th}$ layer based on the correlation measure, 
zeroing out the rest.
Let $I(A, B)$ denote correlation between two set of dimensions:
\begin{equation}
u^L = \argmax_{u} I(u, u^{\text{softmax}})~~\text{s.t.}~~|u^L| = K
\end{equation}
Next, we go to the $(L-1)^\text{th}$ layer, preserving dimensions in the $(L-1)^\text{th}$ layer that are most correlated with  preserved dimensions in the $L^\text{th}$ layer
\begin{equation}
u^{L-1} = \argmax_{u} I(u, u^{L})~~\text{s.t.}~~|u^{L-1}| = K
\end{equation}
This process proceeds until the bottom input embedding layer. An illustration of the proposed layer-wise pruning method is show in Figure \ref{fig:overview}. Algorithm \ref{algo} describes the pruning process.

\begin{algorithm}[t]
\small
  \SetKwInOut{Input}{Input}
  \SetKwInOut{Output}{Output}
  \Input{A trained model $F$ before pruning; the correlation function between two sets of dimensions $I(\cdot,\cdot)$; a specified sparsity $K$; $u^{\text{softmax}}$}
  \Output{Sets of indexes for preserved dimensions $u^1,\cdots,u^L$ in each layer}
  $u^L = \argmax_{u} I(u, u^{\text{softmax}})~~\text{s.t.}~~|u^L| = K$\;
  \tcp*[h]{Top-down layer-wise pruning}
  
  \For{$i\leftarrow L-1$ \KwTo $0$}{ 
    $u^i = \argmax_{u} I(u, u^{i+1})~~\text{s.t.}~~|u^i| = K$\;
  }
  \caption{Layer-wise Pruning}
  \label{algo}
\end{algorithm}

\subsection{Mutual Information between Dimensions} 
Here, we describe quantitative ways to compute  correlation scores $I(A,B)$
between
 dimensions in layer $l-1$ and layer $l$ using MI. 
 \subsubsection{MI for Dimension Selection}
Mutual information (MI) is a measure between two random variables to quantify the amount of information obtained about one  variable through the other variable. 
In our case, we wish to compute the MI between  dimensions $u^l$  at layer $l$ and 
dimensions $u^{l-1}$  at layer $l-1$. 
Let $v_{d^l_k}$ denote the variable for the neuron value of the ${d^l_k}$-th dimension at the $l^\text{th}$ layer. 
MI between $u^l$ and $u^{l-1}$ is given by: 
\begin{equation}
\begin{aligned}
I(u^l,&u^{l-1}) = H(u^l) - H(u^l|u^{l-1})
\end{aligned}
\label{mutual}
\end{equation}
To tangibly compute Eq.(\ref{mutual}), we make assumptions that both $v_{d^l_1},..., v_{d^l_K}$ and $v_{d^{l-1}_1}, ...,v_{d^{l-1}_K}$ are samples from Gaussian distributions:
\begin{equation}
\begin{aligned}
v_{d^l_1},..., v_{d^l_K},v_{d^{l-1}_1}, ...,v_{d^{l-1}_K} \sim &\mathcal{N}(\eta^{l-1,l}_{u^l}, \Sigma^{l-1,l}_{u^l})\\
v_{d^l_1}, ...,v_{d^l_K}  \sim &\mathcal{N}(\eta^{l}_{u^l}, \Sigma^{l}_{u^l})\\
v_{d^{l-1}_1}, ...,v_{d^{l-1}_K}\sim &\mathcal{N}(\eta^{l-1}_{u^l}, \Sigma^{l-1}_{u^l})\\
\end{aligned}
\end{equation}
where  $\eta^{l-1,l}_{u^l}\in\mathbb{R}^{2K\times 1}$; $\eta^{l-1}_{u^l}, \eta^{l}_{u^l}\in\mathbb{R}^{K\times 1}$;
$\Sigma^{l-1,l}_{u^l}\in\mathbb{R}^{2K\times 2K}$; $\Sigma^{l-1}_{u^l}, \Sigma^{l}_{u^l}\in\mathbb{R}^{K\times K}$. 
$\eta$ and $\Sigma$ can be estimated using maximum likelihood. Specifically, for all $(X,Y)\in \mathcal{M}$, we first compute the neuron values for all instances for all layers. 
$\eta^l_{u^l}$ and $\Sigma^l_{u^l}$ are
given  as follows:  
\begin{equation}
\begin{aligned}
\eta^{l}_{u^l} &= \frac{1}{\sum_{X\in \mathcal{M}}|N_x|}\sum_{X\in \mathcal{M}}\sum_{t\in [1,N_x]} {v}_{t,d^l_{{u^l}}} \\
\Sigma^l_{u^l}&= \frac{1}{\sum_{X\in \mathcal{M}}|N_x|}\sum_{X\in \mathcal{M}}\sum_{t\in [1,N_x]} \\
&~~~~~~~~~~~~~~~~~~~~~~ ({v}_{t,d^l_{{u^l}}} -\eta^{l})^\top ({v}_{t,d^l_{{u^l}}} - \eta^{l})
\end{aligned}
\end{equation} 
where ${v}_{t,d^l_{{u^l}}}$ is a vector of length $K$, 
corresponding to a sub-vector within $h^l_t$ with dimension $u^l$. 
$\eta^{l-1}_{u^l}$,  $\eta^{l,l-1}_{u^l}$,  $\Sigma^{l-1}_{u^l}$, $\Sigma^{l,l-1}_{u^l}$ can be computed similarly. 

It is worth noting that 
the proposed model relies on the Gaussian assumption for MI computations, and several recent efforts have been proposed to release this strong assumption, such as
training independent neural nets to estimate MI \cite{belghazi2018mutual}, 
using variational distributions to approximate the distribution \cite{cheng2020club,poole2019variational}. 
These workarounds to avoid the Gaussian assumption requires learning another model  (an independent neural model in \newcite{belghazi2018mutual} and 
variational distributions in \newcite{cheng2020club}) through gradient updates,
and thus cannot
be adapted to the 
 scale in our situation, where  we have to estimate MI for all dimensions across all layers.
The adopted Gaussian model is efficient in estimating MI values in bulk, and achieve satisfying performances.  
We leave how to relax this assumption to future work. 
 
\subsubsection{Greedy Selection}
Selecting $u^l$ based on Eq.(\ref{mutual}) is an NP-hard optimization problem, because the set of possible combinations of dimensions grows exponentially since there are $\mathrm{C}_D^K$ combinations of dimensions ($D$ is the dimension of vector and $K$ is the number of dimensions to pick).
We thus turn to  a greedy forward step-wise selection strategy, a widely used strategy in mutual-information based feature selection. 
Specifically,
let $u^l_{(k)}$ be the set of selected dimensions at time step $k\leq K$. At each time step, we incrementally  add one dimension $d_k^l$ to $u^l_{(k-1)}$
by selecting the dimension that leads to the biggest increase. 
We repeat this process  $K$ times:
\begin{equation}
d_k^l = \argmax_{d\notin u^{l-1}_{(k-1)}} I(u^l,  u^{l-1}_{(k-1)}\cup d)
\label{simple1}
\end{equation}
Inspired by \newcite{brown2012conditional}, 
 further assumptions are made that the selected dimensions are independent and class-conditionally independent given  unselected features, transforming Eq.(\ref{simple1}) to the following form:
 \begin{equation}
 \begin{aligned}
d_k^l &= \argmax_{d\notin u^{l-1}_{(k-1)}}\{I(u^l, d) - \\
&~~~ [\alpha I(d, u^l_{(k-1)}) - \beta I(d, u^l_{(k-1)}|u^l)]\}
\end{aligned}
\label{simple2}
\end{equation}
It is straightforward to see that the first part of Eq.(\ref{simple2}), i.e., $I(u^l, d) $ models the relevance of  selected dimensions,
against the  redundancy compared to the dimensions already selected
, manifested in the second 
and the third part. 
 The model degenerates to 
the model of
Maximum Relevancy Minimum Redundancy (mRMR) \citep{peng2005feature} when $\beta=0$. 
\subsubsection{Squeezing Weights and Features}
For weight matrixes $W$ and feature $H^l$ involved in
the matrix manipulation $W H^l$,  
we do not need to
 actually compute the Hadamard product in Eq\ref{Hadamard}.
 Instead, for $H$, we squeeze all preserved dimensions to the left side and truncate the rest.
For $W$, 
 rows and columns that correspond to pruned dimensions  will be erased and the remaining dimensions will be squeezed. 
 For example, with $m^l=[1,1,0,1]$ and $m^{l+1}=[0,1,1,1]$, the third row and first column of the original matrix $W=[w_{ij}]$ can be pruned, the result of which is  squeezed into a smaller matrix:
\begin{equation}
  \small
  W=\begin{bmatrix}\cancel{w_{11}}&w_{12}&w_{13}&w_{14}\\\cancel{w_{21}}&w_{22}&w_{23}&w_{24}\\\cancel{w_{31}}&\cancel{w_{32}}&\cancel{w_{33}}&\cancel{w_{34}}\\\cancel{w_{41}}&w_{42}&w_{43}&w_{44}\end{bmatrix}
  \Rightarrow \begin{bmatrix}
    w_{12}&w_{13}&w_{14}\\w_{22}&w_{23}&w_{24}\\w_{42}&w_{43}&w_{44}
  \end{bmatrix}
\end{equation}
This avoids irregular memory accesses and thus can significantly speed up matrix-vector product. 
Figure \ref{fig:overview} gives a tangible illustration.

\subsection{Iterative Pruning} 
 Instead of aggressively reducing dimensions from $D$ to $K$ in only one iteration, 
iterative pruning  \cite{han2015learning} gradually reduces model dimensions in multiple steps: 
in each iteration, pruning is followed by model retraining using preserved dimensions. 
As we will show in experiments, this strategy achieves better performances than the single-step pruning with  the same sparsity levels. 

\subsection{Retraining Pruned Dimensions}
The proposed MI based pruning strategy can not only 
be used 
for reducing model size, but also for improving model performances. 
We can view the MI pruning model from a feature selection perspective: given fixed size of features (where we view each neural dimension as a feature), we wish that 
all features in each neural layer be informative and relevant. To this end, we can first remove redundant or irrelevant features,  add new features, retrain the model, and repeat this process.
 This strategy is akin to  feature selection methods in SVMs or logistic regression \cite{kuncheva2007stability}.
 
In the neural  setup, we can achieve this goal by (1) pruning  irrelevant dimensions; (2) reinitializing pruned dimensions (adding new features); and (3) retraining the model. Preserved dimensions and weight matrices are fixed during model retraining,  and we only update pruned dimensions. 
We report the performances of pruning and retraining 60\% dimensions.
It is worth noting that the strategy of retraining pruned dimensions does not serve as the goal of  speedup and model compressing, as pruned dimensions are relearned, making the model of the same size as the model before pruning. 
We as view retraining pruned dimensions as a byproduct of the pruning, with the goal of improving  performances. 
\subsection{Discussions}
For the $Wh$ matrix multiplication in neural models, 
we refer to $W$ as weights, and $h$ as features. 
Weight-based methods  \citep{han2015deep,han2015learning} prune networks based on values of $W$, removing features with smaller weights, 
which are comparable to $L1$ or 
$L2$ regularizers for feature selection \cite{ng2004feature,ravikumar2010high}. 
 MI-based pruning method is comparable to MI based feature selection, which  
attaches attentions to the features by measuring 
feature-label 
correlations \cite{kuncheva2007stability,yu2008stable}. 
\section{Experiments}
We conduct experiments on both generative and discriminative NLP tasks. 
For generative tasks, we conduct experiments on WMT14 En-Fr and
WMT14 En-DE. 
The WMT14 En-Fr dataset consist of 36M and is split into  32000 word-piece
vocabulary. 
The 
WMT 2014 En-DE dataset consisting of about 4.5 million
sentence pairs.  We use BPE \citep{sennrich-etal-2016-neural}  to maintain a source-target vocabulary of 37,000. 
We  use Transformers \citep{vaswani2017attention} as the model backbone.
We use En-Fr to perform comprehensive analysis where 
we use four model setups: extra-large, large, base and tiny.
The model statistics are shown in Table \ref{tab:model}.
It is worth noting that the large and base models are identical to models in \newcite{vaswani2017attention}. 
We train different models  with 16 V100 GPUs with 32G memories.
We follow protocols in \newcite{vaswani2017attention}. 
 Adam \citep{kingma2014adam} is used for all models with $\beta_1$ = 0.9, $\beta_2$ = 0.98 and $\epsilon = 10^{-6}$.
A dropout rate of 0.1 is applied to all layers across all models, and the strategy of label smoothing  \citep{szegedy2016rethinking} is used with smoothing value set to $0.1$.\footnote{Since our goal is to test the performances of different pruning techniques in the vanilla supervised setup, no advanced MT techniques such as backtranslation \citep{sennrich2016back-translation,edunov2018understanding}, self-learning \cite{He2020Revisiting,sun2020neural}, data noising \cite{xie2017data,bengio2015scheduled},
nearest neighbor search \cite{khandelwal2020nearest,meng2021fast,zheng2021adaptive}
 are used.}
 We 
use beam search with a beam size of 20, with no penalty on length. 
We report BLEU scores based on \texttt{multi-bleu.perl} of single models (no ensemble), average floating-point operations (FLOPs), and average practical speedup.

For discriminative tasks, 
we followed the current trend of LM pretraining \cite{devlin2018bert,liu2019roberta,jiao2019tinybert,radford2019language,lan2019albert,brown2020language,clark2020electra,sun2021chinesebert}. 
We test different pruning models on the tasks of question answering  \citep{rajpurkar2016squad,rajpurkar2018know}, 
natural language inference \cite{bowman2015large,williams2017broad} and text classification \cite{socher2013recursive,tang2014learning,howard2018universal,chai2020description,lin2021bertgcn}. 
We use BERT \citep{devlin2018bert} as the backbone, and fine-tune BERT on different datasets. 
Adam \citep{kingma2014adam} is used for all models, with batch size, learning rate and the number of epochs treated as hyper-parameters to be tuned on the dev set. 
 We compare the proposed strategy with the following weight based pruning models:
\begin{itemize}[noitemsep,topsep=0pt,parsep=0pt,partopsep=0pt]
 \item {\it Magnitude Pruning} \citep{han2015learning}: removing weights based on their absolute weight values.
 \item  {\it Movement Pruning} \citep{sanh2020movement}: removing weights
  based on the first-order derivative.
 \item {\it L0 Pruning} \citep{louizos2017learning}:  using the $L_0$  loss to regularize the number of non-zero weights. 
 \end{itemize}

\begin{table}[t]
  \small
  \centering
  \scalebox{0.9}{
  \begin{tabular}{lccccc}\toprule
    {\bf Model} & {\bf $d_\text{model}$} & {\bf $d_\text{ff}$} & {\bf $L$} & {\bf $H$} & {\bf \# Params}\\\midrule
    {\it Extra-Large} & 2,048 & 8,192 & 8 & 16 & 1.1B\\
    {\it Large} & 1,024 & 4,096 & 6 & 16 & 275M\\
    {\it Base} & 512 & 2,048 & 6 & 8 & 93M\\
    {\it Tiny} & 256 & 1,024 & 6 & 8 & 35M \\\bottomrule
  \end{tabular}
  }
  \caption{Model statistics. $d_\text{model}$, $d_\text{ff}$, $L$ and $H$ respectively denote input/output dimensionality, inner-layer dimensionality, \# layers and \# heads.}
  \label{tab:model}
\end{table}

\begin{table}[t]
\center
\small
\scalebox{0.80}{
\begin{tabular}{lcccc}\toprule
{\bf Model} & {\bf BLEU} & {\bf FLOPs} & {\bf Speedup} & {\bf \# Params}\\\midrule
\multicolumn{5}{c}{\underline{\it Original Models}} \\
{\it Extra-Large} & 43.3 & 100$\%$ & 1 & 100$\%$\\
{\it Large} & 41.8 & 24 $\%$ & $\times$ 2.7&$25\%$ \\
{\it Base}& 37.9 &4.2$\%$ & $\times$ 8.6 &$8.5\%$\\
{\it Tiny}& 32.4 &2.3$\%$ & $\times$ 13.7 &$3.2\%$\\
\midrule
\multicolumn{5}{c}{\underline{{\it Without Retraining: Pruning Extra-Large}}} \\
{\it MI} (to large)&42.4 & 22$\%$ & $\times$ 2.6& $25\%$\\
{\it MI} (to base)&39.6 & 4.4$\%$ & $\times$ 8.8& $8.5\%$\\
{\it MI} (to tiny)&34.9 & 2.1$\%$ & $\times$ 13.6& $3.2\%$\\
\cdashline{1-5}[0.8pt/2pt]
{\it Magnitude} (to large)& 41.7 & 23$\%$ & $\times$  2.1& $25\%$\\ 
{\it Magnitude} (to base)& 37.3 & 4.1$\%$ & $\times$  4.5&  $8.5\%$\\ 
{\it Magnitude} (to tiny)& 32.3 & 2.3$\%$ & $\times$  7.5&  $3.2\%$\\ 
\cdashline{1-5}[0.8pt/2pt]
{\it Movement} (to large)& 42.0 & 24$\%$ & $\times$  1.9& $25\%$\\ 
{\it Movement} (to base)& 38.2 & 4.6$\%$ & $\times$  4.7&  $8.5\%$\\ 
{\it Movement} (to tiny)& 33.6 & 2.6$\%$ & $\times$  6.1&  $3.2\%$\\ 
\cdashline{1-5}[0.8pt/2pt]
{\it L0} (to large)& 42.0 & 25$\%$ & $\times$  2.1& $25\%$\\ 
{\it L0} (to base)& 38.0 & 3.9$\%$ & $\times$  3.9&  $8.5\%$\\ 
{\it L0} (to tiny)& 33.8 & 2.3$\%$ & $\times$  5.8&  $3.2\%$\\ 
\midrule
\multicolumn{5}{c}{\underline{{\it Without Retraining: Pruning Large}}} \\
{\it MI} (to base)&38.6 & 4.1$\%$ & $\times$ 8.5& $8.5\%$\\
{\it MI} (to tiny)&33.6& 2.4$\%$ & $\times$ 14.1& $3.2\%$\\\cdashline{1-5}[0.8pt/2pt]

{\it Magnitude} (to base)& 38.3 & 4.5$\%$ & $\times$  4.0&  $8.5\%$\\ 
{\it Magnitude} (to tiny)&32.7& 2.6$\%$ & $\times$ 6.5& $3.2\%$\\\cdashline{1-5}[0.8pt/2pt]
{\it Movement} (to base)& 38.1 & 4.8$\%$ & $\times$  4.7&  $8.5\%$\\ 
{\it Movement} (to tiny)&33.3& 2.4$\%$ & $\times$ 8.3& $3.2\%$\\\cdashline{1-5}[0.8pt/2pt]
{\it L0} (to base)& 38.2 & 4.4$\%$ & $\times$  4.6&  $8.5\%$\\ 
{\it L0} (to tiny)&32.8& 2.9$\%$ & $\times$ 6.9& $3.2\%$\\\midrule

\multicolumn{5}{c}{\underline{{\it Without Retraining: Pruning Base}}} \\
{\it MI} (to tiny)&33.1& 2.3$\%$ & $\times$ 13.5& $3.2\%$\\
{\it Magnitude} (to tiny)&32.5& 2.5$\%$ & $\times$ 8.4& $3.2\%$\\
{\it Movement} (to tiny)&32.8& 2.7$\%$ & $\times$ 8.7& $3.2\%$\\
{\it L0} (to tiny)&32.7& 2.4$\%$ & $\times$ 6.9& $3.2\%$\\\midrule

\multicolumn{5}{c}{\underline{{\it Retraining Pruned Dimensions}}} \\
{\it MI+Extra-Large} & 43.9 (+0.6) & 100$\%$ & 1 & 100$\%$\\
{\it MI+Large} & 42.3 (+0.5) & 24 $\%$ & $\times$ 2.7&$25\%$ \\
{\it MI+Base}& 38.4 (+0.5)  &4.2$\%$ & $\times$ 8.6 &$8.5\%$\\
\bottomrule
\end{tabular}
}
\caption{Test results for WMT14 En-Fr. ``MI'' stands for the propose MI based pruning method, ``Magnitude'' stands for magnitude pruning, ``Movement'' stands for movement pruning and ``L0'' stands for L0 pruning. {\it to X} means pruning the original model to $X$, and $X$ is thus smaller than the original model.
60\% dimensions are pruned and then retrained for the retraining setup. 
}
\label{MT-result}
\end{table}

\begin{table}
\center
\small
\scalebox{0.80}{
\begin{tabular}{lcccc}\toprule
{\bf Model} & {\bf BLEU} & {\bf FLOPs} & {\bf Speedup} & {\bf \# Params}\\\midrule
\multicolumn{5}{c}{\underline{\it Original Models}} \\
{\it Large} & 28.4 & 100 $\%$ & $\times$ 1&$100\%$ \\
{\it Base}& 27.3 &17.5$\%$ & $\times$ 3.1 &$34\%$\\
{\it Tiny}& 23.6 &9.6$\%$ & $\times$ 5.1 &$13\%$\\
\midrule
\multicolumn{5}{c}{\underline{{\it Without Retraining: Pruning Large}}} \\
{\it MI} (to base)&27.9 & 19.2$\%$ & $\times$ 2.6& $34\%$\\
{\it MI} (to tiny)&25.8 & 12.4$\%$ & $\times$ 4.9& $13\%$\\\cdashline{1-5}[0.8pt/2pt]
{\it Magnitude} (to base)& 27.3 & 24.2$\%$ & $\times$  1.8&  $34\%$\\ 
{\it Magnitude} (to tiny)&24.8& 14.1$\%$ & $\times$ 2.8& $13\%$\\\cdashline{1-5}[0.8pt/2pt]
{\it Movement} (to base)&27.6 & 22.4$\%$ & $\times$  1.7&  $34\%$\\ 
{\it Movement} (to tiny)&25.5& 13.0$\%$ & $\times$ 3.5& $13\%$\\\cdashline{1-5}[0.8pt/2pt]
{\it L0} (to base)& 27.6 & 21.9$\%$ & $\times$  1.5&  $34\%$\\ 
{\it L0} (to tiny)&25.8& 13.6$\%$ & $\times$ 2.7& $13\%$\\\midrule

\multicolumn{5}{c}{\underline{{\it Retraining Pruned Dimensions}}} \\
{\it MI+Large} & 28.8 (+0.4) & 17.2 $\%$ & $\times$ 3.2&$34\%$ \\
{\it MI+Base}& 27.9 (+0.6)  &9.8$\%$ & $\times$ 5.0 &$13\%$\\
\bottomrule
\end{tabular}
}
\caption{Test results for WMT14 En-De.}
\label{En-De}
\end{table}

\begin{figure*}[t]
    \centering
    \includegraphics[scale=0.3]{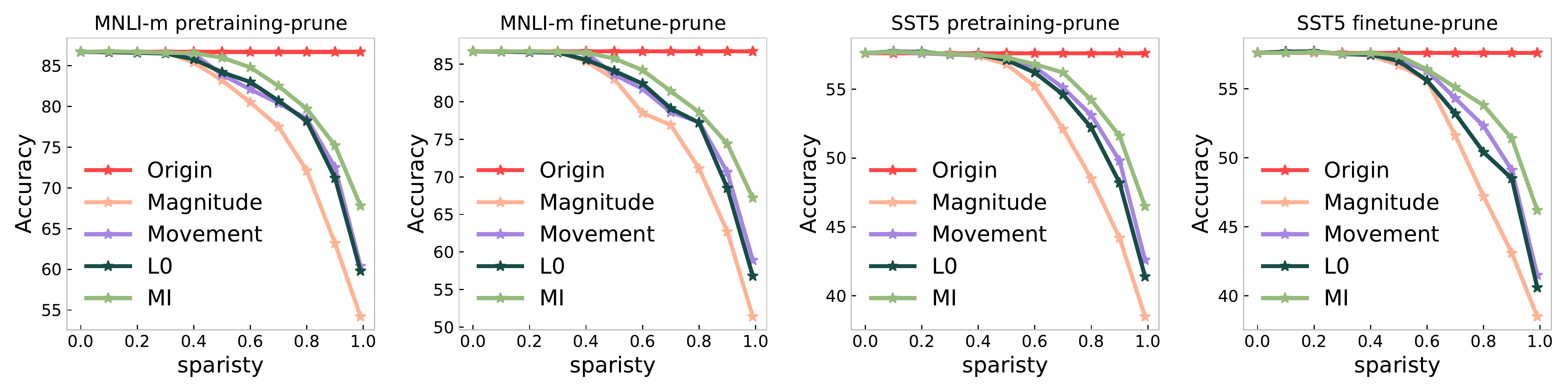}
    \caption{Performances of {\it pretrain-prune} and {\it finetune-prune} on MNLI-m and SST-5.}
    \label{fig:cls}
\end{figure*}

\begin{figure*}[t]
  \centering
  \begin{minipage}[t]{0.5\linewidth}
    \centering
    \includegraphics[scale=0.26]{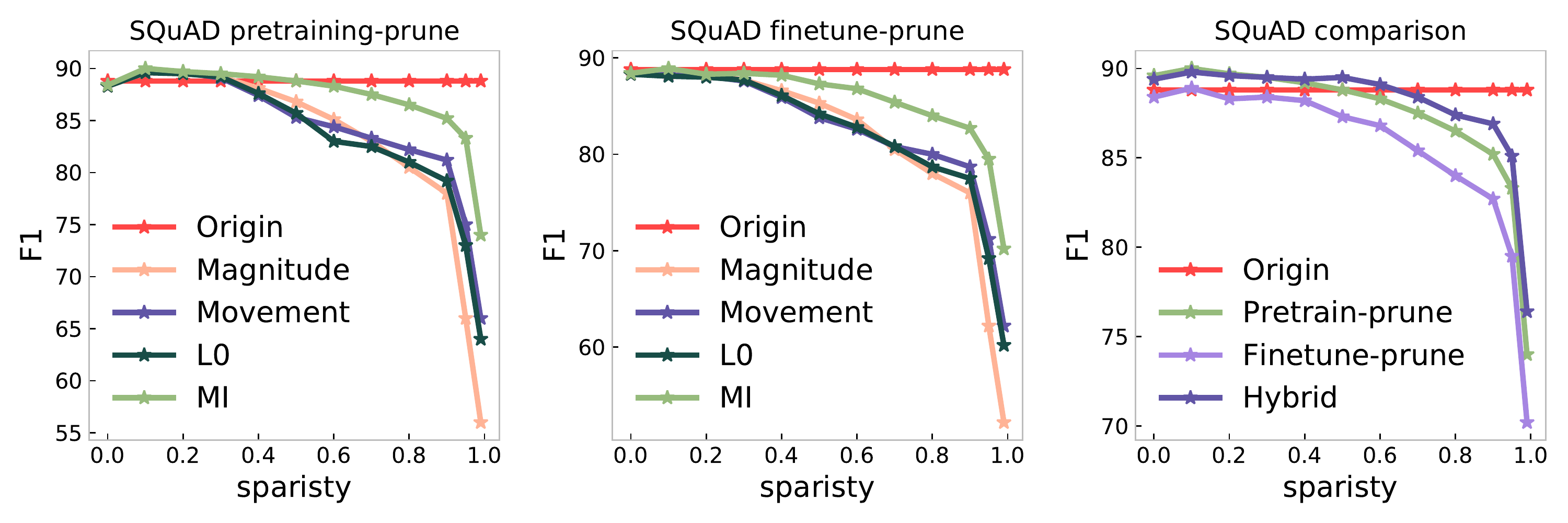}
    \caption{Performances of {\it pretrain-prune}, {\it finetune-prune} and {\it hybrid} on SQuAD.}
    \label{fig:squad}
  \end{minipage}%
  \hfill
  \begin{minipage}[t]{0.5\linewidth}
    \centering
    \includegraphics[scale=0.26]{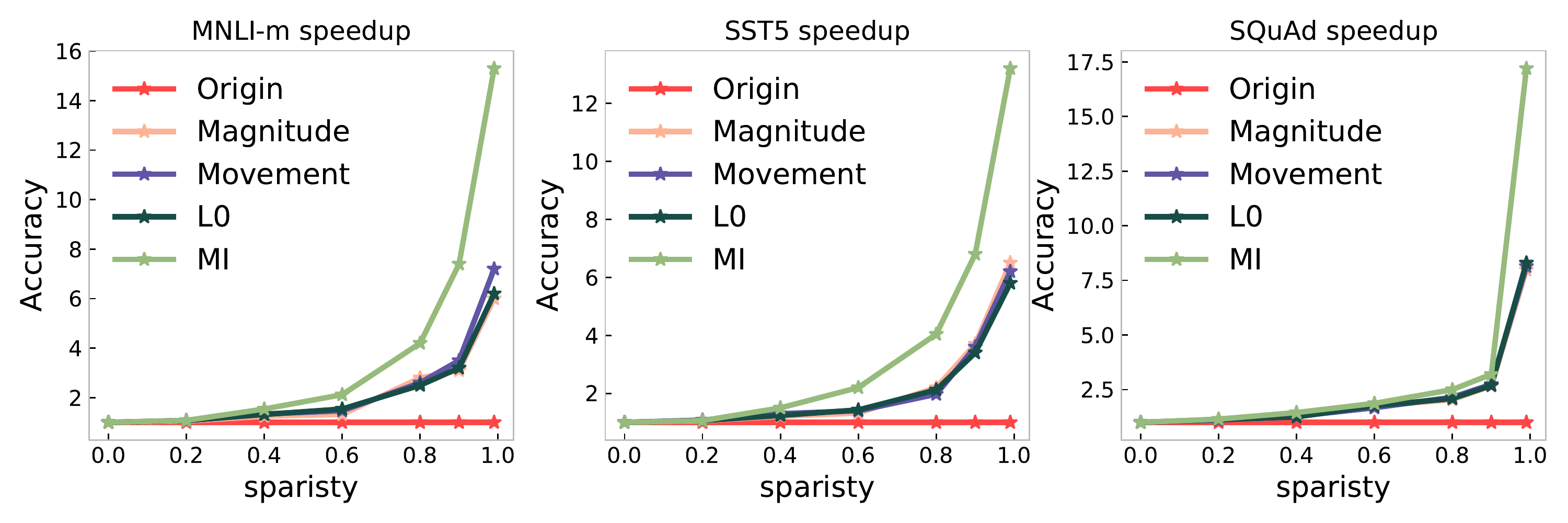}
    \caption{Speedups of different models for {\it pretrain-prune} on MNLI-m, SST-5 and SQuAD.}
    \label{fig:speedup}
  \end{minipage}%
\end{figure*}

\subsection{MT Results}
MT results are shown in Tables \ref{MT-result} and \ref{En-De}. Observations can be summarized  as follows:
(1)  When 
comparing with movement and 
magnitude pruning, at the same levels  of sparsity, the proposed MI method yields greater speedup. 
This is due to the  fact that using MI, the weight matrix $W$ can be squeezed avoiding 
irregular memory accesses.
For  
magnitude and movement pruning: though $W$ is sparse, pruned dimensions in $W$ are scattered and  irregular memory accesses are inevitable. 

(2) 
The MI model yields not only speedup but also performance boosts: 
we find that the proposed MI pruning consistently works better, both in the low-sparsity and high-sparsity situations. 
This is because the mutual information strategy provides a more global feature (dimension) selection strategy based on the output label, rather than focusing on the local matrix weights in matrix manipulations. 
Regarding magnitude pruning  and movement pruning, we find that movement pruning  underperforms magnitude pruning at lower sparsity levels but works better at higher sparsity levels. 

(2) Based on MI, training a big model and then pruning it to a smaller one outperforms directly training a smaller model of the same size, e.g., pruning  extra-large  to large yields a BLEU score of 42.4 for En-Fr, which is +0.6 higher than vanilla large (41.8).
This is also the case with pruning  extra-large to base and tiny, and
pruning large to base and tiny.
The explanations are as follows: a directly trained model contains redundant and  irrelevant  dimensions; 
for the large-training-then-pruning strategy, the model first learns a larger set of feature dimensions, and then prunes irrelevant ones. This makes the model consist of fewer irrelevant feature dimensions than the one directly trained, 
leading to better performances.

(3) Pruning  and then retraining  yields consistent performance boosts over direct training: 
+0.6 for extra-large (43.3 vs 43.9), +0.5 for large (41.8 vs 42.3) and +0.5 for base (37.9 vs 38.4) for En-Fr.
This is because direct training  introduces redundant and less relevant features; 
retraining pruned dimensions  can help the model replace less relevant dimensions with  relevant ones, obtaining 
a state-of-the-art performance of 43.9 BLEU score for En$\rightarrow$Fr translation in setups without back-translation or external data.
Similar phenomenon are observed for En-De with +0.4 for the large model, and +0.6 for base models.


\subsection{BERT Pruning}
We carry out experiments on the pretrained model of BERT-large\footnote{
which contains 24 layers, 1,024 hidden units per layer, 16 heads per layer and 340M parameters in total}. 
We select different degrees of sparsities from 0$\%$ to 90$\%$ at an interval of $10\%$. 
Model pruning can happen either in the pretraining stage ({\it pretrain-prune}), the 
fine-tune stage ({\it finetune-prune}), and both ({\it hybrid}):
 For {\it hybrid}, pruning happens at both stages, with the ultimate sparsity level $\gamma$ being the product of the sparsity level of two stages, $\gamma_{\text{pretrain}}$
and $\gamma_{\text{finetune}}$. 
We compare the performance of the three strategies
on the SQuAD v1.1, MNIL and 
 and SST-5 in Figure \ref{fig:cls} and Figure \ref{fig:squad}. 
Generally, 
{\it pretrain-prune} works consistently better than {\it finetune-prune} with the same level of sparsity.
This is because the  training objective at the pretraining stage is a more general one than that at the finetuning stage, 
with more training data points and categories. 
Pruning at the finetuning stage is  more prone to overfitting, leading to inferior performances. 
The {\it hybrid} method outperforms the {\it pretrain-prune} strategy if the sparsity levels at two stages are carefully calibrated. This is because the {\it hybrid} model 
can progressively prune less relevant dimensions in  pretraining and then less relevant dimensions in task-specific finetuning, leading to better final performances. 

For both {\it pretrain-prune} and {\it finetune-prune}, we find that the proposed MI method offers greater  speedup and better performances  at the same sparsity levels. 
Similar phenomenon are found for MNIL and SST-5.
Figure \ref{fig:speedup} shows the speedup gains for different models for the {\it pretrain-prune} setup. With the same sparsity, random pruning and the proposed MI based pruning lead to the largest speedup, followed by magnitude pruning, movement pruning and $L_0$ pruning. This observation validates that condensed weights serve as an effective remedy for irregular memory access.

\section{Ablation Studies}
In this section, we conduct ablation studies to get a better understanding of model behaviors. 
We use  SQuAD  for analysis, where BERT-large is used.

\subsection{The Effect of $\alpha$ and $\beta$}
\begin{figure}
  \centering
  \includegraphics[scale=0.35]{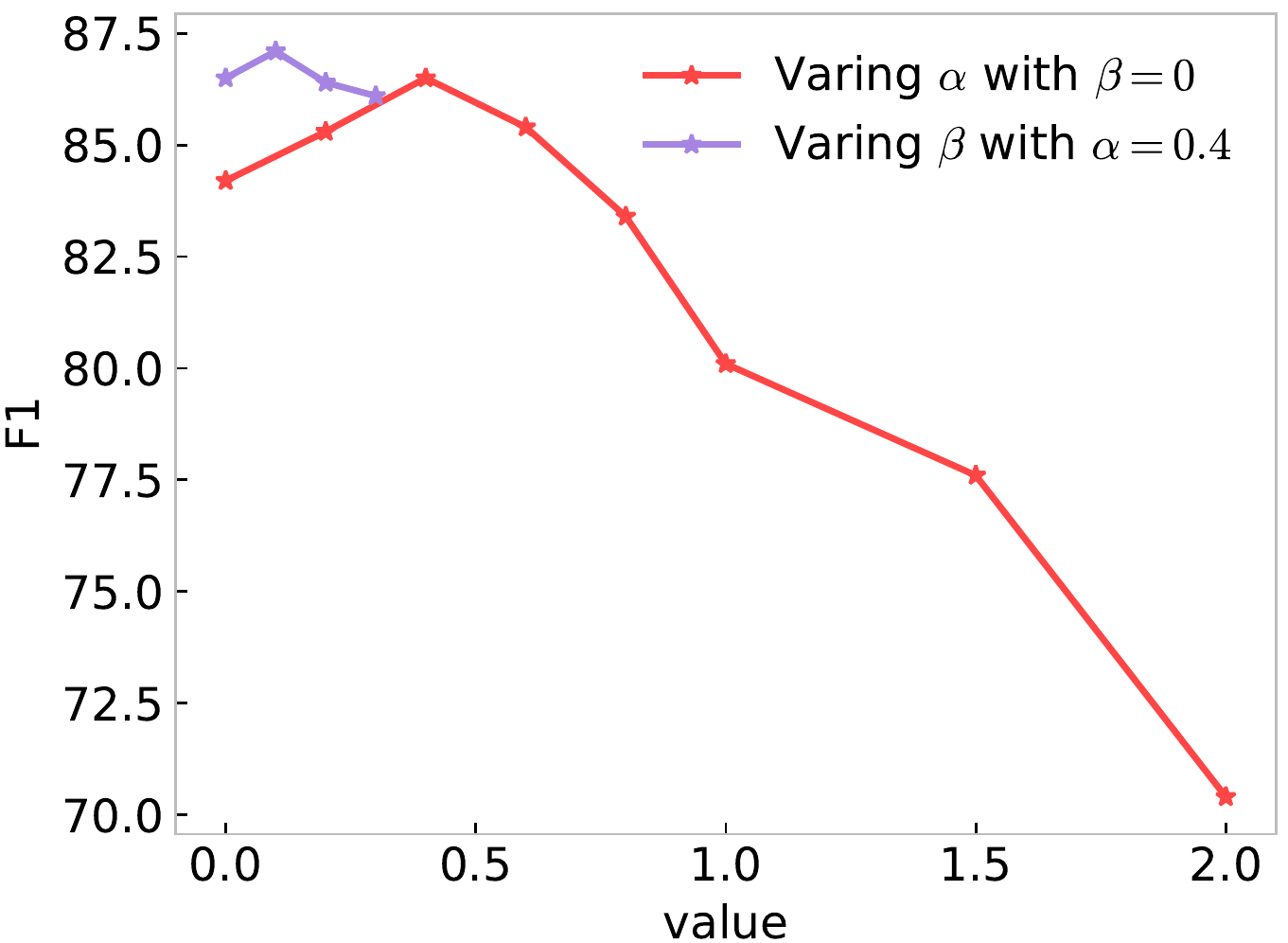}
  \caption{The effect of $\alpha$ and $\beta$.}
  \label{fig:alpha}
\end{figure}

The value of $\alpha$
and $\beta$ in Eq.(\ref{simple2})
 controls the tradeoff between selecting relevant dimensions and 
 removing
 redundant dimensions. 
Based on the {\it pretrain-prune} strategy with sparsity level of 20$\%$, 
we can see from Figure \ref{fig:alpha} that the model works best when the value of $\alpha$ is set to 0.4, and then deteriorates as $\alpha$ increases when fixing $\beta=0$.
With fixed value of $\alpha=0.4$, we find that the influence from  $\beta$ is less significant. This shows that given  
the conditional independency assumption, the improvement from the class-conditionally independent assumption is marginal.
We thus suggest omitting this part if computing resources are limited.  

\subsection{The Effect of Iterative Pruning}
Table \ref{tab:iterative} presents results with different number of pruning iterations, where we use linear interpolation to obtain  sparsity levels for different iterations.
As can be seen, though more pruning iterations lead to better performances, the boost becomes marginal when iteration number exceeds 2. 

\begin{table}[t]
  \small
  \centering
  \begin{tabular}{ccccc}\toprule
    {\it \# Iterations} & 1 & 2 & 3 & 4\\
    {\it F1} & 85.1 & 86.5 & 86.7 & {\bf 86.8}\\
    \bottomrule
  \end{tabular}
  \caption{The effect of iterative pruning.}
  \label{tab:iterative}
\end{table}

\subsection{The Effect of $\gamma_{\text{pretrain}}$
and $\gamma_{\text{finetune}}$}
Fixing the overall sparsity of 0.2,
we explore the effect of $\gamma_{\text{pretrain}}$ and $\gamma_{\text{finetune}}$.
When  $\gamma_{\text{finetune}}=1$, it means we only perform pruning at the pretraining stage;
When  $\gamma_{\text{finetune}}=0.2$, it means we only perform pruning at the finetuning stage. 
As can be seen from Table \ref{tab:gamma}, performance peaks when $\gamma_{\text{finetune}}$ is slightly lower than 1 ($\gamma_{\text{finetune}}=0.8$ , $\gamma_{\text{pretrain}}=0.25$),
and then declines as we increase $\gamma_{\text{finetune}}$. This further validates that 
the final performance benefits more when 
most pruning happens at the pretraining stage. 

\begin{table}[t]
  \small
  \centering
  \begin{tabular}{cccccc}\toprule
    { $\gamma_\text{finetune}$} & 1.0 & 0.8 & 0.6 & 0.4 & 0.2\\
    {\it F1} & 86.5 & {\bf 87.4} & 86.3 & 85.1 & 84.0\\
    \bottomrule
  \end{tabular}
  \caption{The effect of $\gamma_\text{pretrain}$ and $\gamma_\text{finetune}$.}
  \label{tab:gamma}
\end{table}

\begin{table}[t]
  \small
  \centering
  \begin{tabular}{cccc}\toprule
 {\it Method} & Inverted Pyramid & Vanilla &  Pyramid \\
 {\it F1} & {\bf 87.9} & 87.4 & 87.2\\
    \bottomrule
  \end{tabular}
  \caption{Layers with different sparsity values}
  \label{tab:layer}
\end{table}

\subsection{Layers with Different  Sparsity Values}
We explore the situation where given fixed overall sparsity value, different layers can have different levels of sparsity. 
We additionally consider two setups, {\it pyramid}, where lower layers are denser and thus
 less sparse than upper layers, and {\it inverted pyramid}  where upper layers are less sparse than lower layers. 
For {\it pyramid}, with the overall sparsity of 0.2, the lowest word embedding starts with a sparsity level of 0.1, with the sparsity of all layers forms an arithmetic sequence. 
 {\it inverted pyramid} has the same overall sparsity value of 0.2, with the lowest word embedding starts with a sparsity level of 0.3.
Results are shown in Table \ref{tab:layer}. We can observe that  {\it inverted pyramid} outperforms 
{\it vanilla}, which outperforms  {\it  pyramid}.
These results illustrate that to obtain better performances in model pruning with fixed overall sparsity, upper layers should be less sparse than lower layers.
This is because upper layers contain more high-level and dense information about the input. Therefore, pruning upper layers does more harm to the model. 
 Lower layers contain more noise, and thus hurt the model less when get pruned. 

\section{Conclusion and Future Work}
In this paper, we propose MI based methods for model pruning in NLP.
The proposed model avoids the issue of irregular memory access, leading to 
higher speedup with the same level of sparsity. Also, the proposed strategy prunes the model in a top-down fashion based on global training signals, and thus achieves higher accuracies. 
In future work, we should release the strong assumption that neuron values come from a Gaussian distribution. 

\bibliography{emnlp2020}
\bibliographystyle{acl_natbib}

\end{document}